\documentclass[twoside,11pt]{article}
\pdfoutput=1

\usepackage{geometry}
\usepackage{natbib}
\usepackage{graphicx}
\usepackage{rotating}
\usepackage{array}
\usepackage{algorithm}
\usepackage{algorithmic}
\usepackage{amsmath}
\usepackage{amssymb}

\usepackage{theorem}
\newcommand{\BlackBox}{\rule{1.5ex}{1.5ex}}  

\newcommand{\cbr}[1]{\left\{#1\right\}}

\newcommand{\captsize}{\fontsize{10.5pt}{\baselineskip}\selectfont}

\newcommand{\E}{\mathbb{E}}
\newcommand{\KL}{\text{KL}}
\newcolumntype{L}[1]{>{\raggedright\arraybackslash}p{#1}}
\newcolumntype{R}[1]{>{\raggedleft\arraybackslash}p{#1}}
\newcolumntype{C}[1]{>{\centering\let\newline\\\arraybackslash\hspace{0pt}}m{#1}}
\newcolumntype{?}{!{\vrule width 1pt}}
\makeatletter
\newcommand{\thickhline}{%
    \noalign {\ifnum 0=`}\fi \hrule height 1pt
    \futurelet \reserved@a \@xhline
}
\newcolumntype{"}{@{\hskip\tabcolsep\vrule width 1pt\hskip\tabcolsep}}
\makeatother

\newcommand{\intset}[1]{\cbr{1..n}}

\usepackage[colorlinks,pdfpagelabels,plainpages=false]{hyperref}
\usepackage{rotating}

\usepackage{xcolor}
\definecolor{dark-red}{rgb}{0.4,0.15,0.15}
\definecolor{dark-blue}{rgb}{0.15,0.15,0.4}
\definecolor{medium-blue}{rgb}{0,0,0.5}
\hypersetup{
   colorlinks, linkcolor={dark-blue},
   citecolor={dark-blue}, urlcolor={medium-blue}
}

\usepackage{booktabs}
\usepackage{bm}
\usepackage{amsmath}
\usepackage{amssymb}
\usepackage{amsfonts}
\usepackage{mathtools}
\usepackage{comment}
\usepackage{subfigure}

\newcommand{\mbf}[1]{{\boldsymbol{\mathbf{#1}}}}
\renewcommand{\bm}{\mbf}

\usepackage{color}

\usepackage{parskip}

\usepackage{multirow}
\usepackage{titlefoot}
\usepackage{gensymb}

\title{Harnessing Deep Neural Networks with Logic Rules}

\renewcommand{\footnotesize}{\normalsize}

\author{
  Zhiting Hu \\
  \texttt{zhitingh@cs.cmu.edu}
  \and
  Xuezhe Ma \\
  \texttt{xuezhem@cs.cmu.edu}
  \and
  Zhengzhong Liu  \\
  \texttt{liu@cs.cmu.edu}
  \and
  Eduard Hovy \\
  \texttt{hovy@cmu.edu}
  \and
  Eric P. Xing \\
  \texttt{epxing@cs.cmu.edu}
  \and
  {\small School of Computer Science, Carnegie Mellon University}
}

\begin{document}
\date{}

\maketitle

\begin{abstract}

\begin{sloppypar}
Combining deep neural networks with structured logic rules is desirable to harness flexibility and reduce uninterpretability of the neural models. We propose a general framework capable of enhancing various types of neural networks (e.g., CNNs and RNNs) with declarative first-order logic rules. Specifically, we develop an iterative distillation method that transfers the structured information of logic rules into the weights of neural networks.
We deploy the framework on a CNN for sentiment analysis, and an RNN for named entity recognition. With a few highly intuitive rules, we obtain substantial improvements and achieve state-of-the-art or comparable results to previous best-performing systems.
\end{sloppypar}

\end{abstract}

\section{Introduction}
Deep neural networks provide a powerful mechanism for learning patterns from massive data, achieving new levels of performance on image classification~\citep{krizhevsky2012imagenet}, speech recognition~\citep{hinton2012deep}, machine translation~\citep{bahdanau2014neural}, playing strategic board games~\citep{silver2016mastering}, and so forth.

Despite the impressive advances, the widely-used DNN methods still have limitations. The high predictive accuracy has heavily relied on large amounts of labeled data; and the purely data-driven learning can lead to uninterpretable and sometimes counter-intuitive results~\citep{szegedy2013intriguing,nguyen2015deep}. It is also difficult to encode human intention to guide the models to capture desired patterns, without expensive direct supervision or ad-hoc initialization.

On the other hand, the cognitive process of human beings have indicated that people learn not only from concrete examples (as DNNs do) but also from different forms of general knowledge and rich experiences \citep{minsky1980learning,lake2015human}.
{\it Logic rules} provide a flexible declarative language for communicating high-level cognition and expressing structured knowledge. It is therefore desirable to integrate logic rules into DNNs, to transfer human intention and domain knowledge to neural models, and regulate the learning process.

In this paper, we present a framework capable of enhancing general types of neural networks, such as convolutional networks (CNNs) and recurrent networks (RNNs), on various tasks, with logic rule knowledge. Combining symbolic representations with neural methods have been considered in different contexts. Neural-symbolic systems~\citep{garcez2012neural} construct a network from a given rule set to execute reasoning. To exploit {\it a priori} knowledge in general neural architectures, recent work augments each raw data instance with useful features~\citep{collobert2011natural}, while network training, however, is still limited to instance-label supervision and suffers from the same issues mentioned above. Besides, a large variety of structural knowledge cannot be naturally encoded in the feature-label form.

Our framework enables a neural network to learn simultaneously from labeled instances as well as logic rules,
through an {\it iterative rule knowledge distillation} procedure that transfers the structured information encoded in the logic rules into the network parameters.
Since the general logic rules are complementary to the specific data labels, a natural ``side-product'' of the integration is the support for semi-supervised learning where unlabeled data is used to better absorb the logical knowledge. Methodologically, our approach can be seen as a combination of the knowledge distillation~\citep{hinton2015distilling,buciluǎ2006model}
and the posterior regularization (PR) method~\citep{ganchev2010posterior}. In particular, at each iteration we adapt the posterior constraint principle from PR to construct a rule-regularized {\it teacher}, and train the {\it student} network of interest to
imitate the predictions of the teacher network.
We leverage soft logic to support flexible rule encoding.

We apply the proposed framework on both CNN and RNN, and deploy on the task of sentiment analysis (SA) and named entity recognition (NER), respectively. With only a few (one or two) very intuitive rules, both the distilled networks and the joint teacher networks strongly improve over their basic forms (without rules), and achieve better or comparable performance to state-of-the-art models which typically have more parameters and complicated architectures.

To the best of our knowledge, this is the first work to integrate logic rules with general workhorse types of deep neural networks in a principled framework. The encouraging results indicate our method can be potentially useful for incorporating richer types of human knowledge, and improving other application domains.

\section{Related Work}
Combination of logic rules and neural networks has been considered in different contexts.
Neural-symbolic systems~\citep{garcez2012neural}, such as KBANN~\citep{towell1990refinement} and CILP++~\citep{francca2014fast}, construct network architectures from given rules to perform reasoning and knowledge acquisition. A related line of research, such as Markov logic networks~\citep{richardson2006markov}, derives probabilistic graphical models (rather than neural networks) from the rule set.

With the recent success of deep neural networks in a vast variety of application domains,
it is increasingly desirable to incorporate structured logic knowledge into general types of networks to harness flexibility and reduce uninterpretability. Recent work that trains on extra features from domain knowledge \citep{collobert2011natural}, while producing improved results, does not go beyond the data-label paradigm. \citet{kulkarni2015deep} uses a specialized training procedure with careful ordering of training instances to obtain an interpretable neural layer of an image network. \citet{karaletsos2016bayesian} develops a generative model jointly over data-labels and similarity knowledge expressed in triplet format to learn improved disentangled representations.

Though there do exist general frameworks that allow encoding various structured constraints on latent variable models~\citep{ganchev2010posterior,zhu2014bayesian,liang2009learning}, they either are not directly applicable to the NN case, or could yield inferior performance as in our empirical study. \citet{liang2008structure} transfers predictive power of pre-trained structured models to unstructured ones in a pipelined fashion.

Our proposed approach is distinct in that we use an iterative rule distillation process to effectively transfer rich structured knowledge, expressed in the declarative first-order logic language, into parameters of general neural networks. We show that the proposed approach strongly outperforms an extensive array of other either ad-hoc or general integration methods.

\section{Method}
\begin{figure}[!htp]
\centering
\includegraphics[scale=0.53]{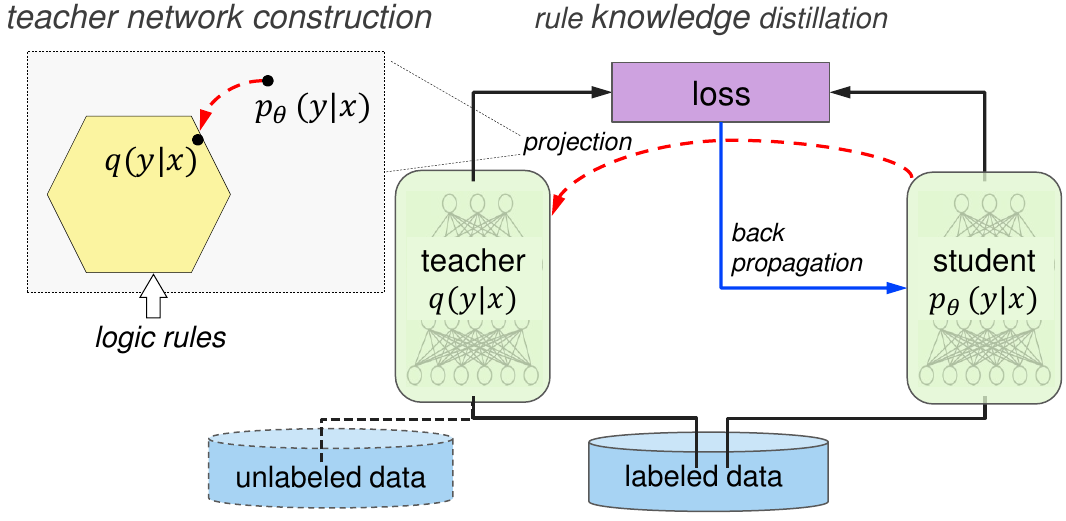}
\caption{Framework Overview. At each iteration, the teacher network is obtained by projecting the student network to a rule-regularized subspace (red dashed arrow); and the student network is updated to balance between emulating the teacher's output and predicting the true labels (black/blue solid arrows).}
\label{fig:arch}
\end{figure}

In this section we present our framework which encapsulates the logical structured knowledge into a neural network. This is achieved by forcing the network to emulate the predictions of a rule-regularized teacher, and evolving both models iteratively throughout training (section~\ref{sec:distill}). The process is agnostic to the network architecture, and thus applicable to general types of neural models including CNNs and RNNs. We construct the teacher network in each iteration by adapting the posterior regularization principle in our logical constraint setting (section~\ref{sec:teacher}), where our formulation provides a closed-form solution. Figure~\ref{fig:arch} shows an overview of the proposed framework.

\subsection{Learning Resources: Instances and Rules}
Our approach allows neural networks to learn from both specific examples and general rules. Here we give the settings of these ``learning resources''.

Assume we have input variable $\bm{x}\in\mathcal{X}$ and target variable $\bm{y}\in\mathcal{Y}$. For clarity, we focus on $K$-way classification, where $\mathcal{Y}=\Delta^K$ is the $K$-dimensional probability simplex and $\bm{y}\in\{0,1\}^K\subset\mathcal{Y}$ is a one-hot encoding of the class label. However, our method specification can straightforwardly be applied to other contexts such as regression and sequence learning (e.g., NER tagging, which is a sequence of classification decisions). The training data $\mathcal{D}=\{(\bm{x}_n,\bm{y}_n)\}_{n=1}^{N}$ is a set of instantiations of $(\bm{x}, \bm{y})$.

Further consider a set of first-order logic (FOL) rules with confidences, denoted as $\mathcal{R}=\{(R_{l}, \lambda_{l})\}_{l=1}^{L}$, where $R_l$ is the $l$th rule over the input-target space $(\mathcal{X}, \mathcal{Y})$, and $\lambda_l\in[0,\infty]$ is the confidence level with $\lambda_l=\infty$ indicating a hard rule, i.e., all groundings are required to be true (=1). Here a grounding is the logic expression with all variables being instantiated.
Given a set of examples $(\bm{X}, \bm{Y})\subset(\mathcal{X},\mathcal{Y})$ (e.g., a minibatch from $\mathcal{D}$), the set of groundings of $R_l$ are denoted as $\{r_{lg}(\bm{X},\bm{Y})\}_{g=1}^{G_l}$.
In practice a rule grounding is typically relevant to only a single or subset of examples, though here we give the most general form on the entire set.

We encode the FOL rules using soft logic~\citep{bach2015hinge} for flexible encoding and stable optimization. Specifically, soft logic allows continuous truth values from the interval $[0,1]$ instead of $\{0,1\}$, and the Boolean logic operators are reformulated as:
\begin{equation}\label{eq:psl}
\small
\begin{split}
& A\&B =\max\{A+B-1,0\} \\
& A\vee B =\min\{A+B,1\} \\
& A_1 \wedge \dots \wedge A_N = \sum\nolimits_{i}A_i / N \\
& \neg A = 1 - A
\end{split}
\end{equation}
Here $\&$ and $\wedge$ are two different approximations to logical conjunction~\citep{foulds2015latent}: $\&$ is useful as a selection operator (e.g., $A\&B=B$ when $A=1$, and $A\&B=0$ when $A=0$), while $\wedge$ is an averaging operator.

\subsection{Rule Knowledge Distillation}
\label{sec:distill}
A neural network defines a conditional probability $p_{\theta}(\bm{y}|\bm{x})$ by using a softmax output layer that produces a $K$-dimensional soft prediction vector denoted as $\bm{\sigma}_{\theta}(\bm{x})$. The network is parameterized by weights $\bm{\theta}$.
Standard neural network training has been to iteratively update $\bm{\theta}$ to produce the correct labels of training instances.
To integrate the information encoded in the rules,
we propose to train the network to also imitate the outputs of a rule-regularized projection of $p_{\theta}(\bm{y}|\bm{x})$, denoted as $q(\bm{y}|\bm{x})$, which explicitly includes rule constraints as regularization terms. In each iteration $q$ is constructed by projecting $p_\theta$ into a subspace constrained by the rules, and thus has desirable properties.
We present the construction in the next section. The prediction behavior of $q$ reveals the information of the regularized subspace and structured rules. Emulating the $q$ outputs serves to transfer this knowledge into $p_{\theta}$. The new objective is then formulated as a balancing between imitating the soft predictions of $q$ and predicting the true hard labels:
\begin{equation}~\label{eq:distill}
\small
\begin{split}
\bm{\theta}^{(t+1)} = \arg\min_{\theta\in\Theta} \frac{1}{N}\sum_{n=1}^{N}
&(1-\pi) \ell(\bm{y}_n, \bm{\sigma}_{\theta}(\bm{x}_n)) \\
& + \pi \ell(\bm{s}^{(t)}_n, \bm{\sigma}_{\theta}(\bm{x}_n)),
\end{split}
\end{equation}
where $\ell$ denotes the loss function selected according to specific applications (e.g., the cross entropy loss for classification); $\bm{s}^{(t)}_n$ is the soft prediction vector of $q$ on $\bm{x}_n$ at iteration $t$; and $\pi$ is the imitation parameter calibrating the relative importance of the two objectives.

A similar imitation procedure has been used in other settings such as model compression~\citep{buciluǎ2006model,hinton2015distilling} where the process is termed {\it distillation}. Following them we call $p_{\theta}(\bm{y}|\bm{x})$ the ``student'' and $q({\bm{y}|\bm{x}})$ the ``teacher'', which can be intuitively explained in analogous to human education where a teacher is aware of systematic general rules and she instructs students by providing her solutions to particular questions (i.e., the soft predictions). An important difference from previous distillation work, where the teacher is obtained beforehand and the student is trained thereafter, is that our teacher and student are learned simultaneously during training.

Though it is possible to combine a neural network with rule constraints by  projecting the network to the rule-regularized subspace after it is fully trained as before with only data-label instances,
or by optimizing projected network directly,
we found our iterative teacher-student distillation approach provides a much superior performance, as shown in the experiments.
Moreover, since $p_{\theta}$ distills the rule information into the weights $\bm{\theta}$ instead of relying on explicit rule representations, we can use $p_{\theta}$ for predicting new examples at test time when the rule assessment is expensive or even unavailable (i.e., the {\it privileged information} setting~\citep{lopez2015unifying}) while still enjoying the benefit of integration.
Besides, the second loss term in Eq.\eqref{eq:distill} can be augmented with rich unlabeled data in addition to the labeled examples, which enables {\it semi-supervised} learning for better absorbing the rule knowledge.

\subsection{Teacher Network Construction}
\label{sec:teacher}
We now proceed to construct the teacher network $q(\bm{y}|\bm{x})$ at each iteration from $p_{\theta}(\bm{y}|\bm{x})$. The iteration index $t$ is omitted for clarity. We adapt the posterior regularization principle in our logic constraint setting. Our formulation ensures a closed-form solution for $q$ and thus avoids any significant increases in computational overhead.

Recall the set of FOL rules $\mathcal{R}=\{(R_{l}, \lambda_{l})\}_{l=1}^{L}$.
Our goal is to find the optimal $q$ that fits the rules while at the same time staying close to $p_{\theta}$. For the first property, we apply a commonly-used strategy that imposes the rule constraints on $q$ through an expectation operator. That is, for each rule (indexed by $l$) and each of its groundings (indexed by $g$) on  $(\bm{X},\bm{Y})$, we expect $\E_{q(\bm{Y}|\bm{X})}[r_{lg}(\bm{X},\bm{Y})]=1$, with confidence $\lambda_l$. The constraints define a rule-regularized space of all valid distributions. For the second property, we measure the closeness between $q$ and $p_{\theta}$ with KL-divergence, and wish to minimize it. Combining the two factors together and further allowing slackness for the constraints, we finally get the following optimization problem:
\begin{equation}~\label{eq:pr}
\small
\begin{split}
\min_{q, \bm{\xi}\geq 0}\ \KL &(q(\bm{Y}|\bm{X}) \| p_{\theta}(\bm{Y}|\bm{X})) + C\sum\nolimits_{l,g_l}\xi_{l,g_l} \\
\text{s.t.}\ \  &\lambda_{l}(1-\E_{q}[r_{l,g_l}(\bm{X},\bm{Y})]) \leq \xi_{l,g_l} \\
&\ g_l = 1, \dots, G_l,\ \  l = 1, \dots, L,
\end{split}
\end{equation}
where $\xi_{l,g_l}\geq 0$ is the slack variable for respective logic constraint; and $C$ is the regularization parameter. The problem can be seen as projecting $p_{\theta}$ into the constrained subspace.
The problem is convex and can be efficiently solved in its dual form with closed-form solutions. We provide the detailed derivation in the supplementary materials and directly give the solution here:
\begin{equation}~\label{eq:q}
\small
\begin{split}
q^*(\bm{Y} | \bm{X}) \propto p_{\theta}(\bm{Y}|\bm{X}) \exp\left\{-\sum_{l,g_l}C\lambda_{l}(1-r_{l,g_l}(\bm{X},\bm{Y}))\right\}
\end{split}
\end{equation}
Intuitively, a strong rule with large $\lambda_l$ will lead to low probabilities of predictions that fail to meet the constraints. We discuss the computation of the normalization factor in section~\ref{sec:implement}.

Our framework is related to the posterior regularization (PR) method \citep{ganchev2010posterior} which places constraints over model posterior in unsupervised setting.
In classification, our optimization procedure is analogous to the modified EM algorithm for PR, by using cross-entropy loss in Eq.\eqref{eq:distill} and evaluating the second loss term on unlabeled data differing from $\mathcal{D}$, so that Eq.\eqref{eq:q} corresponds to the E-step and Eq.\eqref{eq:distill} is analogous to the M-step. This sheds light from another perspective on why our framework would work. However, we found in our experiments (section~\ref{sec:exp}) that to produce strong performance it is crucial to use the same labeled data $\bm{x}_n$ in the two losses of Eq.\eqref{eq:distill} so as to form a direct trade-off between imitating soft predictions and predicting correct hard labels.

\subsection{Implementations}
\label{sec:implement}
The procedure of iterative distilling optimization of our framework is summarized in Algorithm~\ref{alg:opt}.

During training we need to compute the soft predictions of $q$ at each iteration, which is straightforward through direct enumeration if the rule constraints in Eq.\eqref{eq:q} are factored in the same way as the base neural model $p_{\theta}$ (e.g., the ``but''-rule of sentiment classification in section~\ref{sec:sent}). If the constraints introduce additional dependencies, e.g., bi-gram dependency as the transition rule in the NER task (section~\ref{sec:ner}), we can use dynamic programming for efficient computation. For higher-order constraints (e.g., the listing rule in NER), we approximate through Gibbs sampling that iteratively samples from $q(\bm{y}_i|\bm{y}_{-i},\bm{x})$ for each position $i$. If the constraints span multiple instances, we group the relevant instances in minibatches for joint inference (and randomly break some dependencies when a group is too large).
Note that calculating the soft predictions is efficient since only one NN forward pass is required to compute the base distribution $p_\theta(\bm{y}|\bm{x})$ (and few more, if needed, for calculating the truth values of relevant rules).

\paragraph{$p$ v.s. $q$ at Test Time}
At test time we can use either the distilled student network $p$, or the teacher network $q$ after a final projection. Our empirical results show that both models substantially improve over the base network that is trained with only data-label instances. In general $q$ performs better than $p$. Particularly, $q$ is more suitable when the logic rules introduce additional dependencies (e.g., spanning over multiple examples), requiring joint inference.
In contrast, as mentioned above, $p$ is more lightweight and efficient, and useful when rule evaluation is expensive or impossible at prediction time. Our experiments compare the performance of $p$ and $q$ extensively.

\paragraph{Imitation Strength $\pi$}
The imitation parameter $\pi$ in Eq.\eqref{eq:distill} balances between emulating the teacher soft predictions and predicting the true hard labels. Since the teacher network is constructed from $p_{\theta}$, which, at the beginning of training, would produce low-quality predictions, we thus favor predicting the true labels more at initial stage. As training goes on, we gradually bias towards emulating the teacher predictions to effectively distill the structured knowledge. Specifically, we define $\pi^{(t)}=\min\{\pi_0, 1-\alpha^{t}\}$ at iteration $t\geq 0$, where $\alpha \leq 1$ specifies the speed of decay and $\pi_0 < 1$ is a lower bound.

\begin{algorithm}[t]
\small
\centering
\caption{\small Harnessing NN with Rules}
\label{alg:opt}
\begin{algorithmic}[1]
\REQUIRE The training data $\mathcal{D}=\{(\bm{x}_{n}, \bm{y}_{n})\}_{n=1}^{N}$,\\
\quad\ \  The rule set $\mathcal{R} = \{(R_l, \lambda_l)\}_{l=1}^{L}$, \\
\quad\ \  Parameters: $\pi$ -- imitation parameter \\
\qquad\qquad\qquad\ \ $C$ -- regularization strength
\STATE Initialize neural network parameter $\bm{\theta}$
\REPEAT
	\STATE Sample a minibatch $(\bm{X},\bm{Y}) \subset \mathcal{D}$
    \STATE Construct teacher network $q$ with Eq.\eqref{eq:q}
    \STATE Transfer knowledge into $p_{\theta}$ by updating $\bm{\theta}$ with Eq.\eqref{eq:distill}
\UNTIL{convergence}
\ENSURE Distill student network $p_{\theta}$ and teacher network $q$
\end{algorithmic}
\end{algorithm}

\section{Applications}
We have presented our framework that is general enough to improve various types of neural networks with rules, and easy to use in that users are allowed to impose their knowledge and intentions through the declarative first-order logic.
In this section we illustrate the versatility of our approach by applying it on two workhorse network architectures, i.e., convolutional network and recurrent network, on two representative applications, i.e., {\it sentence-level sentiment analysis} which is a classification problem, and {\it named entity recognition} which is a sequence learning problem.

For each task, we first briefly describe the base neural network. Since we are not focusing on tuning network architectures, we largely use the same or similar networks to previous successful neural models. We then design the linguistically-motivated rules to be integrated.

\subsection{Sentiment Classification}
\label{sec:sent}
Sentence-level sentiment analysis is to identify the sentiment (e.g., positive or negative) underlying an individual sentence. The task is crucial for many opinion mining applications. One challenging point of the task is to capture the contrastive sense (e.g., by conjunction ``but'') within a sentence.

\begin{figure}[t]
\hspace{0.3in}
  \subfigure
  {\includegraphics[width=0.42\textwidth]{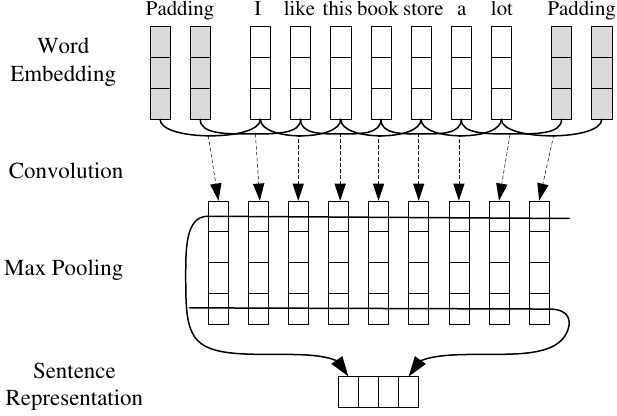}}
\hspace{0.5in}
  \subfigure
  {\includegraphics[width=0.33\textwidth]{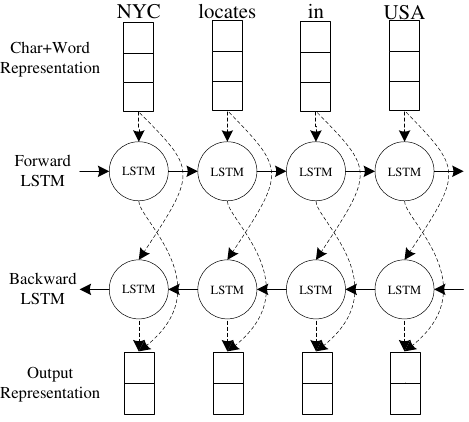}}
  \caption{{\bf Left}: The CNN architecture for sentence-level sentiment analysis. The sentence representation vector is followed by a fully-connected layer with softmax output activation, to output sentiment predictions.
  {\bf Right}: The architecture of the bidirectional LSTM recurrent network for NER. The CNN for extracting character representation is omitted.}
\label{fig:base-nn}
\end{figure}

\paragraph{Base Network}
We use the single-channel convolutional network proposed in \citep{kim2014convolutional}. The simple model has achieved compelling performance on various sentiment classification benchmarks. The network contains a convolutional layer on top of word vectors of a given sentence, followed by a max-over-time pooling layer and then a fully-connected layer with softmax output activation. A convolution operation is to apply a filter to word windows. Multiple filters with varying window sizes are used to obtain multiple features. Figure~\ref{fig:base-nn}, left panel, shows the network architecture.

\paragraph{Logic Rules}
One difficulty for the plain neural network is to identify contrastive sense in order to capture the dominant sentiment precisely. The conjunction word ``but'' is one of the strong indicators for such sentiment changes in a sentence, where the sentiment of clauses following ``but'' generally dominates.
We thus consider sentences $S$ with an ``A-but-B'' structure, and expect the sentiment of the whole sentence to be consistent with the sentiment of clause $B$. The logic rule is written as:
\begin{equation}\label{eq:rule-but}
\small
\begin{split}
&\text{has-`A-but-B'-structure}(S) \Rightarrow \\
&\left( \bm{1}(y=+) \Rightarrow \bm{\sigma}_{\theta}(B)_{+}\ \wedge\ \bm{\sigma}_{\theta}(B)_{+} \Rightarrow \bm{1}(y=+) \right),
\end{split}
\end{equation}
where $\bm{1}(\cdot)$ is an indicator function that takes 1 when its argument is true, and 0 otherwise; class `+' represents `positive'; and $\bm{\sigma}_{\theta}(B)_{+}$ is the element of $\bm{\sigma}_{\theta}(B)$ for class '+'. By Eq.\eqref{eq:psl}, when $S$ has the `A-but-B' structure, the truth value of the above logic rule equals to $(1+\bm{\sigma}_{\theta}(B)_{+})/2$ when $y=+$, and $(2-\bm{\sigma}_{\theta}(B)_{+})/2$ otherwise~\footnote{\footnotesize Replacing $\wedge$ with $\&$ in Eq.\eqref{eq:rule-but} leads to a probably more intuitive rule which takes the value $\bm{\sigma}_{\theta}(B)_{+}$ when $y=+$, and $1-\bm{\sigma}_{\theta}(B)_{+}$ otherwise.}. Note that here we assume two-way classification (i.e., positive and negative), though it is straightforward to design rules for finer grained sentiment classification.

\subsection{Named Entity Recognition}
\label{sec:ner}
NER is to locate and classify elements in text into entity categories such as ``persons'' and ``organizations''.
It is an essential first step for downstream language understanding applications. The task assigns to each word a named entity tag in an ``X-Y'' format where X is one of BIEOS (Beginning, Inside, End, Outside, and Singleton) and Y is the entity category. A valid tag sequence has to follow certain constraints by the definition of the tagging scheme. Besides, text with structures (e.g., lists) within or across sentences can usually expose some consistency patterns.

\paragraph{Base Network}
The base network has a similar architecture with the bi-directional LSTM recurrent network (called BLSTM-CNN) proposed in \citep{chiu2015named} for NER which has outperformed most of previous neural models.
The model uses a CNN and pre-trained word vectors to capture character- and word-level information, respectively. These features are then fed into a bi-directional RNN with LSTM units for sequence tagging.  Compared to \citep{chiu2015named} we omit the character type and capitalization features, as well as the additive transition matrix in the output layer. Figure~\ref{fig:base-nn}, right panel, shows the network architecture.

\paragraph{Logic Rules}
The base network largely makes independent tagging decisions at each position, ignoring the constraints on successive labels for a valid tag sequence (e.g., I-ORG cannot follow B-PER). In contrast to recent work~\citep{lample2016neural}  which adds a conditional random field (CRF) to capture bi-gram dependencies between outputs, we instead apply logic rules which does not introduce extra parameters to learn. An example rule is:
\begin{equation}\label{eq:rule-trans}
\small
\text{equal}(y_{i-1}, \text{I-ORG}) \Rightarrow \neg\ \text{equal}(y_{i}, \text{B-PER})
\end{equation}
The confidence levels are set to $\infty$ to prevent any violation.

We further leverage the {\it list} structures within and across sentences of the same documents. Specifically, named entities at corresponding positions in a list are likely to be in the same categories. For instance, in ``1. Juventus, 2. Barcelona, 3. ...'' we know ``Barcelona'' must be an organization rather than a location, since its counterpart entity ``Juventus'' is an organization. We describe our simple procedure for identifying lists and counterparts in the supplementary materials. The logic rule is encoded as:
\begin{equation}\label{eq:rule-list}
\small
\text{is-counterpart}(X,A) \Rightarrow 1 - \| c(\bm{e}_{y}) - c(\bm{\sigma}_{\theta}(A)) \|_2,
\end{equation}
where $\bm{e}_{y}$ is the one-hot encoding of $y$ (the class prediction of $X$); $c(\cdot)$ collapses the probability mass on the labels with the same categories into a single probability, yielding a vector with length equaling to the number of categories. We use $\ell_2$ distance as a measure for the closeness between predictions of $X$ and its counterpart $A$. Note that the distance takes value in $[0,1]$ which is a proper soft truth value. The list rule can span multiple sentences (within the same document).
We found the teacher network $q$ that enables explicit joint inference provides much better performance over the distilled student network $p$ (section~\ref{sec:exp}).

\section{Experiments}
\label{sec:exp}
We validate our framework by evaluating its applications of sentiment classification and named entity recognition on a variety of public benchmarks. By integrating the simple yet effective rules with the base networks, we obtain substantial improvements on both tasks and achieve state-of-the-art or comparable results to previous best-performing systems. Comparison with a diverse set of other rule integration methods demonstrates the unique effectiveness of our framework. Our approach also shows promising potentials in the semi-supervised learning and sparse data context.

Throughout the experiments we set the regularization parameter to $C=6$. In sentiment classification we set the imitation parameter to $\pi^{(t)}=1-0.95^{t}$, while in NER $\pi^{(t)}=\min\{0.9, 1-0.9^{t}\}$ to downplay the noisy listing rule.
The confidence levels of rules are set to $\lambda_l=1$, except for hard constraints whose confidence is $\infty$.
For neural network configuration, we largely followed the reference work, as specified in the following respective sections.
All experiments were performed on a Linux machine with eight 4.0GHz CPU cores, one Tesla K40c GPU, and 32GB RAM. We implemented neural networks using Theano~\footnote{http://deeplearning.net/software/theano}, a popular deep learning platform.

\subsection{Sentiment Classification}
\subsubsection{Setup}
We test our method on a number of commonly used benchmarks, including {\bf 1) SST2}, Stanford Sentiment Treebank \citep{socher2013recursive} which contains 2 classes (negative and positive), and 6920/872/1821 sentences in the train/dev/test sets respectively. Following \citep{kim2014convolutional} we train models on both sentences and phrases since all labels are provided. {\bf 2) MR} \citep{pang2005seeing}, a set of 10,662 one-sentence movie reviews with negative or positive sentiment. {\bf 3) CR} \citep{hu2004mining}, customer reviews of various products, containing 2 classes and 3,775 instances. For MR and CR, we use 10-fold cross validation as in previous work. In each of the three datasets, around 15\% sentences contains the word ``but''.

For the base neural network we use the ``non-static'' version in \citep{kim2014convolutional} with the exact same configurations. Specifically, word vectors are initialized using word2vec \citep{mikolov2013distributed} and fine-tuned throughout training, and the neural parameters are trained using SGD with the Adadelta update rule \citep{zeiler2012adadelta}.

\begin{table*}[t]
  \centering
   \small
\begin{tabular}{ r  L{200pt}  L{40pt}  L{60pt}  L{55pt}} \cmidrule[\heavyrulewidth]{1-5}
& Model  & SST2 & MR & CR \\ \cmidrule{1-5}
1 & CNN~\citep{kim2014convolutional} & 87.2 & 81.3$\pm$0.1 & 84.3$\pm$0.2  \\
2 & CNN-Rule-$p$ & 88.8 &  81.6$\pm$0.1 & 85.0$\pm$0.3 \\
3 & CNN-Rule-$q$ & 89.3 &  {\bf 81.7$\pm$0.1} & {\bf 85.3$\pm$0.3} \\ \cmidrule{1-5}
4 & MGNC-CNN~\citep{zhang2016mgnc} & 88.4 & -- & -- \\
5 & MVCNN~\citep{yin2015multichannel} & {\bf 89.4} & -- & -- \\
6 & CNN-multichannel~\citep{kim2014convolutional}  & 88.1 & 81.1 & 85.0 \\
7 & Paragraph-Vec~\citep{le2014distributed} & 87.8 & -- & -- \\
8 & CRF-PR~\citep{yang2014context} & -- & -- & 82.7 \\
9 & RNTN~\citep{socher2013recursive} & 85.4 & -- & -- \\
10 & G-Dropout~\citep{wang2013fast} & -- & 79.0 & 82.1 \\ \cmidrule[\heavyrulewidth]{1-5}
\end{tabular}
\caption{Accuracy (\%) of Sentiment Classification. Row 1, CNN~\citep{kim2014convolutional} is the base network corresponding to the ``CNN-non-static'' model in \citep{kim2014convolutional}. Rows 2-3 are the networks enhanced by our framework: CNN-Rule-$p$ is the student network and CNN-Rule-$q$ is the teacher network. For MR and CR, we report the average accuracy$\pm$one standard deviation using 10-fold cross validation.}
\label{tab:sent}
\end{table*}

\subsubsection{Results}
Table~\ref{tab:sent} shows the sentiment classification performance. Rows 1-3 compare the base neural model with the models enhanced by our framework with the ``but''-rule (Eq.\eqref{eq:rule-but}). We see that our method provides a strong boost on accuracy over all three datasets. The teacher network $q$ further improves over the student network $p$, though the student network is more widely applicable in certain contexts as discussed in sections~\ref{sec:distill} and~\ref{sec:implement}.
Rows 4-10 show the accuracy of recent top-performing methods. On the MR and CR datasets, our model outperforms all the baselines. On SST2,  MVCNN~\citep{yin2015multichannel} (Row 5) is the only system that shows a slightly better result than ours. Their neural network has combined diverse sets of pre-trained word embeddings (while we use only word2vec) and contained more neural layers and parameters than our model.

To further investigate the effectiveness of our framework in integrating structured rule knowledge, we compare with an extensive array of other possible integration approaches. Table~\ref{tab:sent-internal} lists these methods and their performance on the SST2 task. We see that: 1) Although all methods lead to different degrees of improvement, our framework outperforms all other competitors with a large margin. 2) In particular, compared to the pipelined method in Row 6 which is in analogous to the structure compilation work~\citep{liang2008structure}, our iterative distillation (section~\ref{sec:distill}) provides better performance. Another advantage of our method is that we only train one set of neural parameters, as opposed to two separate sets as in the pipelined approach.
3) The distilled student network ``-Rule-$p$'' achieves much superior accuracy compared to the base CNN, as well as ``-project''  and ``-opt-project'' which explicitly project CNN to the rule-constrained subspace. This validates that our distillation procedure transfers the structured knowledge into the neural parameters effectively. The inferior accuracy of ``-opt-project'' can be partially attributed to the poor performance of its neural network part which achieves only 85.1\% accuracy and leads to inaccurate evaluation of the ``but''-rule in Eq.\eqref{eq:rule-but}.

\begin{table}[t]
  \centering
  \small
\begin{tabular}{ r  l  l } \cmidrule[\heavyrulewidth]{1-3}
& Model  & Accuracy (\%) \\ \cmidrule{1-3}
1 & CNN~\citep{kim2014convolutional} & 87.2  \\
2 & -but-clause & 87.3 \\
3 & -$\ell_2$-reg & 87.5 \\
4 & -project & 87.9 \\
5 & -opt-project & 88.3 \\
6 & -pipeline & 87.9 \\ \cmidrule{1-3}
7 & -Rule-$p$ & 88.8 \\
8 & -Rule-$q$ & {\bf 89.3} \\ \cmidrule[\heavyrulewidth]{1-3}
\end{tabular}
 \caption{\captsize Performance of different rule integration methods on SST2. 1) CNN is the base network;
 2) ``-but-clause'' takes the clause after ``but'' as input;
 3) ``-$\ell_2$-reg'' imposes a regularization term $\gamma\|\bm{\sigma}_{\theta}(S)-\bm{\sigma}_{\theta}(Y)\|_2$ to the CNN objective, with the strength $\gamma$ selected on dev set; 4) ``-project'' projects the trained base CNN to the rule-regularized subspace with Eq.\eqref{eq:pr}; 5) ``-opt-project'' directly optimizes the projected CNN; 6) ``-pipeline'' distills the pre-trained ``-opt-project'' to a plain CNN; 7-8) ``-Rule-$p$'' and ``-Rule-$q$'' are our models with $p$ being the distilled student network and $q$ the teacher network. Note that ``-but-clause'' and ``-$\ell_2$-reg'' are ad-hoc methods applicable specifically to the ``but''-rule.}
\label{tab:sent-internal}
\end{table}
\begin{table}[t]
  \centering
  \small
\begin{tabular}{r l l l l l} \cmidrule[\heavyrulewidth]{1-6}
& Data size & 5\% & 10\% & 30\% & 100\% \\ \cmidrule{1-6}
1 & CNN & 79.9 & 81.6 & 83.6 & 87.2 \\
2 & -Rule-$p$ & 81.5 & 83.2 & 84.5 & 88.8 \\
3 & -Rule-$q$ & 82.5 & 83.9 & 85.6 & {\bf 89.3} \\ \cmidrule{1-6}
4 & -semi-PR  & 81.5  & 83.1 & 84.6 & -- \\
5 & -semi-Rule-$p$ & 81.7 & 83.3 & 84.7 & -- \\
6 & -semi-Rule-$q$ & {\bf 82.7} & {\bf 84.2} & {\bf 85.7} & -- \\  \cmidrule[\heavyrulewidth]{1-6}
\end{tabular}
\caption{\captsize Accuracy (\%) on SST2 with varying sizes of labeled data and semi-supervised learning. The header row is the percentage of labeled examples for training. Rows 1-3 use only the supervised data. Rows 4-6 use semi-supervised learning where the remaining training data are used as unlabeled examples. For ``-semi-PR'' we only report its projected solution (in analogous to $q$) which performs better than the non-projected one (in analogous to $p$).}
\label{tab:sent-semi}
\end{table}

We next explore the performance of our framework with varying numbers of labeled instances as well as the effect of exploiting unlabeled data. Intuitively, with less labeled examples we expect the general rules would contribute more to the performance, and unlabeled data should help better learn from the rules. This can be a useful property especially when data are sparse and labels are expensive to obtain.
Table~\ref{tab:sent-semi} shows the results. The subsampling is conducted on the sentence level. That is, for instance, in ``5\%'' we first selected 5\% training sentences uniformly at random, then trained the models on these sentences as well as their phrases. The results verify our expectations.
1) Rows 1-3 give the accuracy of using only data-label subsets for training. In every setting our methods consistently outperform the base CNN. 2) ``-Rule-$q$'' provides higher improvement on 5\% data (with margin 2.6\%) than on larger data (e.g., 2.3\% on 10\% data, and 2.0\% on 30\% data), showing promising potential in the sparse data context. 3) By adding unlabeled instances for semi-supervised learning as in Rows 5-6, we get further improved accuracy.
4) Row 4, ``-semi-PR'' is the posterior regularization~\citep{ganchev2010posterior} which imposes the rule constraint through only unlabeled data during training. Our distillation framework consistently provides substantially better results.

\subsection{Named Entity Recognition}
\subsubsection{Setup}
We evaluate on the well-established CoNLL-2003 NER benchmark~\citep{tjong2003introduction}, which contains 14,987/3,466/3,684 sentences and 204,567/51,578/46,666 tokens in train/dev/test sets, respectively. The dataset includes 4 categories, i.e., {\it person, location, organization}, and {\it misc}. BIOES tagging scheme is used. Around 1.7\% named entities occur in lists.

We use the mostly same configurations for the base BLSTM network as in \citep{chiu2015named}, except that, besides the slight architecture difference (section~\ref{sec:ner}), we apply Adadelta for parameter updating. GloVe~\citep{pennington2014glove} word vectors are used to initialize word features.
\begin{table}[t]
  \centering
  \fontsize{8.6pt}{0.9em}\selectfont
\begin{tabular}{@{}r  l  l@{}} \cmidrule[\heavyrulewidth]{1-3}
& Model  & F1 \\ \cmidrule{1-3}
1 & BLSTM & 89.55  \\
2 & BLSTM-Rule-trans  &  $p$: 89.80,\ $q$: 91.11 \\
3 & BLSTM-Rules  &  $p$: 89.93,\ $q$: {\bf 91.18} \\  \cmidrule{1-3}
4 & NN-lex~\citep{collobert2011natural} & 89.59 \\
5 & S-LSTM~\citep{lample2016neural} & 90.33 \\
6 & BLSTM-lex~\citep{chiu2015named} & 90.77 \\
7 & BLSTM-CRF$_1$~\citep{lample2016neural} & 90.94 \\
8 & Joint-NER-EL~\citep{luo2016joint} & 91.20 \\
9 & BLSTM-CRF$_2$~\citep{ma2016end} & {\bf 91.21} \\
\cmidrule[\heavyrulewidth]{1-3}
\end{tabular}
\caption{\captsize Performance of NER on CoNLL-2003. Row 2, BLSTM-Rule-trans imposes the transition rules (Eq.\eqref{eq:rule-trans}) on the base BLSTM. Row 3, BLSTM-Rules further incorporates the list rule (Eq.\eqref{eq:rule-list}). We report the performance of both the student model $p$ and the teacher model $q$.
}
\label{tab:ner}
\end{table}
\subsubsection{Results}
Table~\ref{tab:ner} presents the performance on the NER task. By incorporating the bi-gram transition rules (Row 2), the joint teacher model $q$ achieves 1.56 improvement in F1 score that outperforms most previous neural based methods (Rows 4-7), including the BLSTM-CRF model~\citep{lample2016neural} which applies a conditional random field (CRF) on top of a BLSTM in order to capture the transition patterns and encourage valid sequences. In contrast, our method implements the desired constraints in a more straightforward way by using the declarative logic rule language, and at the same time does not introduce extra model parameters to learn. Further integration of the list rule (Row 3) provides a second boost in performance, achieving an F1 score very close to the best-performing systems including Joint-NER-EL~\citep{luo2016joint} (Row 8), a probabilistic graphical model optimizing NER and entity linking jointly with massive external resources, and BLSTM-CRF~\citep{ma2016end}, a combination of BLSTM and CRF with more parameters than our rule-enhanced neural networks.

From the table we see that the accuracy gap between the joint teacher model $q$ and the distilled student $p$ is relatively larger than in the sentiment classification task (Table~\ref{tab:sent}). This is because in the NER task we have used logic rules that introduce extra dependencies between adjacent tag positions as well as multiple instances, making the explicit joint inference of $q$ useful for fulfilling these structured constraints.

\section{Discussion and Future Work}
We have developed a framework which combines deep neural networks with first-order logic rules to allow integrating human knowledge and intentions into the neural models. In particular, we proposed an iterative distillation procedure that transfers the structured information of logic rules into the weights of neural networks. The transferring is done via a teacher network constructed  using the posterior regularization principle. Our framework is general and applicable to various types of neural architectures. With a few intuitive rules, our framework significantly improves base networks on sentiment analysis and named entity recognition, demonstrating the practical significance of our approach.

Though we have focused on first-order logic rules, we leveraged soft logic formulation which can be easily extended to general probabilistic models for expressing structured distributions and performing inference and reasoning \citep{lake2015human}. We plan to explore these diverse knowledge representations to guide the DNN learning. The proposed iterative distillation procedure also reveals connections to recent neural autoencoders~\citep{kingma2013auto,rezende2014stochastic} where generative models encode probabilistic structures and neural recognition models distill the information through iterative optimization~\citep{rezende2016one,johnson2016structured,karaletsos2016bayesian}.

The encouraging empirical results indicate a strong potential of our approach for improving other application domains such as vision tasks, which we plan to explore in the future.
Finally, we also would like to generalize our framework to automatically learn the confidence of different rules, and derive new rules from data.

\section*{Acknowledgments}
We thank the anonymous reviewers for their valuable comments. This work is supported by NSF IIS1218282, NSF IIS1447676, Air Force FA8721-05-C-0003, and FA8750-12-2-0342.

\newpage
{
\bibliographystyle{apalike}
\bibliography{acl2016}
}

\clearpage{}\newpage{}
\appendix

\numberwithin{equation}{section}

\section{Appendix}

\subsection{Solving Problem Eq.\eqref{eq:pr}, Section~\ref{sec:teacher}}

We provide the detailed derivation for solving the problem in Eq.\eqref{eq:pr}, Section~\ref{sec:teacher}, which we repeat here:
\begin{equation}~\label{supp:eq:pr}
\begin{split}
\min_{q, \bm{\xi}\geq 0}\ \KL &(q(\bm{Y}|\bm{X}) \| p_{\theta}(\bm{Y}|\bm{X})) + C\sum_{l,g_l}\xi_{l,g_l} \\
\text{s.t.}\ \  &\lambda_{l}(1-\E_{q}[r_{l,g_l}(\bm{X},\bm{Y})]) \leq \xi_{l,g_l} \\
&\ g_l = 1, \dots, G_l,\ \  l = 1, \dots, L,
\end{split}
\end{equation}

The following derivation is largely adapted from \citep{ganchev2010posterior} for the logic rule constraint setting, with some reformulation that produces closed-form solution.

The Lagrangian is
\begin{equation}\label{eq:prob-dual}
\begin{split}
\max_{\bm{\mu}\geq 0, \bm{\eta}\geq 0, \alpha\geq 0} \min_{q(\bm{y}), \bm{\xi}} L,
\end{split}
\end{equation}
where
\begin{equation}
\begin{split}
&L = \KL( q(\bm{Y}|\bm{X}) \| p_{\theta}(\bm{Y}|\bm{X})) + \sum_{l,g_l}(C-\mu_{l,g_l})\xi_{l,g_l} \\
&\qquad\qquad\qquad + \sum_{l,g_l}\eta_{l,g_l}\left( \E_q[\lambda_{l} (1-r_{l,g_l}(\bm{X},\bm{Y}))] - \xi_{l,g_l} \right) + \alpha(\sum_{\bm{Y}}q(\bm{Y}|\bm{X}) - 1)
\end{split}
\end{equation}
Solving Eq.\eqref{eq:prob-dual}, we obtain
\begin{equation}\label{supp:eq:sol-q}
\begin{split}
\nabla_{q} L &= \log q(\bm{Y}|\bm{X}) + 1 - \log p_\theta(\bm{Y}|\bm{X}) + \sum_{l,g_l}\eta_{l,g_l}\left[ \lambda_l (1-r_{l,g_l}(\bm{X},\bm{Y})) \right] + \alpha = 0\\
&\qquad\qquad\Longrightarrow\qquad q(\bm{Y}|\bm{X}) = \frac{p_\theta(\bm{Y}|\bm{X}) \exp\left\{-\sum_{l}\eta_{l}\lambda_{l}(1-r_{l,g_l}(\bm{X},\bm{Y}))\right\}}{e\exp(\alpha)}
\end{split}
\end{equation}

\begin{equation}\label{supp:eq:sol-mu}
\begin{split}
\nabla_{\xi_{l,g_l}} L = C - \mu_{l,g_l} - \eta_{l,g_l} = 0\qquad\Longrightarrow\qquad \mu_{l,g_l} = C - \eta_{l,g_l} 
\end{split}
\end{equation}

\begin{equation}\label{supp:eq:sol-alpha}
\begin{split}
\nabla_{\alpha} L &= \sum_{\bm{Y}} \frac{p_\theta(\bm{Y}|\bm{X}) \exp\left\{-\sum_{l,g_l}\eta_{l,g_l}\lambda_{l}(1-r_{l,g_l}(\bm{X},\bm{Y}))\right\}}{e\exp(\alpha)} - 1 = 0 \\
&\qquad\qquad\Longrightarrow\qquad \alpha = \log \left( \frac{\sum_{\bm{Y}}p(\bm{Y} | \bm{X}) \exp\left\{-\sum_{l,g_l}\eta_{l,g_l}\lambda_{l}(1-r_{l,g_l}(\bm{X},\bm{Y}))\right\}}{e} \right)
\end{split}
\end{equation}

Let $Z_{\eta} = \sum_{\bm{Y}}p(\bm{Y}|\bm{X}) \exp\left\{-\sum_{l,g_l}\eta_{l,g_l}\lambda_{l}(1-r_{l,g_l}(\bm{X},\bm{Y}))\right\}$. Plugging $\alpha$ into $L$
\begin{equation}
\begin{split}
L &= -\log Z_{\eta} + \sum_{l,g_l} (C+\mu_{l,g_l})\xi_{l,g_l} - \sum_{l,g_l} \eta_{l,g_l}\xi_{l,g_l} \\
&= -\log Z_{\eta}
\end{split}
\end{equation}
Since $Z_\eta$ monotonically decreases as $\eta$ increases, and from Eq.\eqref{supp:eq:sol-mu} we have $\eta_{l,g_l}\leq C$, therefore:
\begin{equation}\label{supp:eq:sol-eta}
\begin{split}
&\max_{C\geq \eta \geq 0} -\log Z_{\eta}  \\
&\qquad\qquad\Longrightarrow\qquad \eta^{*}_{l,g_l} = C
\end{split}
\end{equation}

Plugging Eqs.\eqref{supp:eq:sol-alpha} and \eqref{supp:eq:sol-eta} into Eq.\eqref{supp:eq:sol-q} we obtain the solution of $q$ as in Eq.(4).

\subsection{Identifying Lists for NER}

We design a simple pattern-matching based method to identify lists and counterparts in the NER task. We ensure high precision and do not expect high recall. In particular, we only retrieve lists that with the pattern ``1. ... 2. ... 3. ...'' (i.e., indexed by numbers), and ``- ... - ... - ...'' (i.e., each item marked with ``-''). We require at least 3 items to form a list.

We further require the text of each item follows certain patterns to ensure the text is highly likely to be named entities, and rule out those lists whose item text is largely free text. Specifically, we require 1) all words of the item text all start with capital letters; 2) referring the text between punctuations as ``block'', each block includes no more than 3 words.

We detect both intra-sentence lists and inter-sentence lists in documents. We found the above patterns are effective to identify true lists. A better list detection method is expected to further improve our NER results.

\end{document}